# Hypergraph $p$-Laplacian Regularization for Remote Sensing Image Recognition

Xueqi Ma, Weifeng Liu, *Senior Member, IEEE*, Shuying Li, and Yicong Zhou, *Senior Member, IEEE*

*Abstract*—It is of great importance to preserve locality and similarity information in semi-supervised learning (SSL) based applications. Graph based SSL and manifold regularization based SSL including Laplacian regularization (LapR) and Hypergraph Laplacian regularization (HLapR) are representative SSL methods and have achieved prominent performance by exploiting the relationship of sample distribution. However, it is still a great challenge to exactly explore and exploit the local structure of the data distribution. In this paper, we present an effect and effective approximation algorithm of Hypergraph $p$-Laplacian and then propose Hypergraph $p$-Laplacian regularization (HpLapR) to preserve the geometry of the probability distribution. In particular, $p$-Laplacian is a nonlinear generalization of the standard graph Laplacian and Hypergraph is a generalization of a standard graph. Therefore, the proposed HpLapR provides more potential to exploiting the local structure preserving. We apply HpLapR to logistic regression and conduct the implementations for remote sensing image recognition. We compare the proposed HpLapR to several popular manifold regularization based SSL methods including LapR, HLapR and HpLapR on UC-Merced dataset. The experimental results demonstrate the superiority of the proposed HpLapR.

*Index Terms*—hypergraph, manifold learning, $p$-Laplacian, remote sensing, semi-supervised learning

## I. INTRODUCTION

The classification of remote sensing images [9] [11] has become an important branch of data mining owing to the speedy development of space technology. However, in practical applications, annotating images is costly and time consuming, so only a small number of labeled samples are available whereas a lot of unlabeled samples are easy to collect. Semi-supervised learning which can make use of labeled and unlabeled data has been investigated to solve this problem. One successful work is manifold regularization, which has attracted considerable attention due to its rich theoretical studies [1] [2] [3] and its excellent performance in multimedia data (e.g., text, image, video, audio, etc.) processing [4] [5] [6] [7] [8] [10]. The main idea is to explore the geometry of the intrinsic data probability distribution to leverage the learning performance. Another is graph based SSL [12] [13], which construct a

Xueqi Ma and Weifeng Liu are with the College of Information and Control Engineering, China University of Petroleum (East China), Qingdao 266580, China. e-mail: liuwf@upc.edu.cn.

Shuying Li is with The 16th Institute, China Aerospace Science and Technology Corporation, Xi'an 710100, China. email: angle_lisy@163.com.

Yicong Zhou is with the Faculty of Science and Technology, University of Macau, Macau, China. email: yicongzhou@umac.mo.

similarity graph over data to exploit the local geometry of both labeled and unlabeled data and have achieved appealing due to its flexibility and low computation complexity in practice.

Manifold regularization (MR) framework [2] exploits the geometry of the probability distribution that generates the data and incorporates it as a regularization term. Laplacian regularization is one prominent manifold regularization based SSL algorithm, which determines the underlying manifold by using the graph Laplacian. Wang *et al.* [15] presented a manifold regularized multi-view subspace clustering (MRMSC) method to better incorporate the correlated and complementary information from different views. The graph Laplacian is constructed to maintain the data manifold locally of each view. Luo *et al.* [5] employed manifold regularization to smooth the functions along the data manifold for multitask learning. Jiang *et al.* [4] presented a muti-manifold method for recognition by exploring the local geometric structure of samples. Liu *et al.* [16] proposed multiview Hessian regularized logistic regression (mHLR) which combining multiple Hessian regularization to leverage the exploring of local geometry. Lu *et al.* [17] built a model of sparse feature selection-based manifold regularization (SFSMR) to select the optimal information and preserve the underlying manifold structure of data for scene recognition.

Typically, in graph based SSL, it is assumed that there is a graph over the data lying on data manifolds. In the graph, vertices represent samples and edge weights indicate the similarity between samples. For example, Zhou *et al.* [14] constructed a directed graph learned from labeled and unlabeled data for web categorization, in which each vertex represents a web page, and each edge represents a hyperlink between two web pages. For graph based SSL, it is essential to construct an effective graph over data with complex distribution. Compared with existing simple graph only models the pairwise relationship of images, Hypergraph learning using a hyperedge to link multiple samples can model the high-order relationship of samples.

In [18], the hypergraph idea is first introduced to the field of computer vision, which is a generalization of a simple graph. Unlike a simple graph that take account of the relationship between two vertices, a set of vertices is connected by a hyperedge in a hypergraph. Thus, the hypergraph contains more local grouping information in comparison to simple graph. Hypergraph has been widely used to image classification [24], ranking [21] [33] and video segmentation [23]. Sun *et al.* [20] constructed a hypergraph to exploit the correlation information among different labels for multi-label learning. Zass *et al.* [19] presented a hypergraph based image matching problem in a probabilistic setting represented by a convex optimization problem. Huang *et al.* [22] proposed a hypergraph based transductive algorithm to the field of image retrieval. Yu *et al.* [24] proposed an adaptive hypergraph learning method for transductive image classification.

In this paper, we propose a Hypergraph $p$-Laplacian regularized method for remote sensing image recognition. The hypergraph and $p$-Laplacian [32] [34] [37] [38] both provide convincing theoretical evidence to better preserve the local structure of data. However, the computation of hypergraph $p$-Laplacian is a strenuous task. We provide an effect and efficient fully approximation algorithm of Hypergraph $p$-Laplacian. Considering the higher order relationship of samples, we build the Hypergraph $p$-Laplacian regularizer for local structure preserving. We introduce Hypergraph $p$-Laplacian regularization (HpLapR) to logistic regression for remote sensing image recognition. We conduct experiments on the UC-Merced data set [25] by comparing with the popular algorithms including Laplacian regularization (LapR), Hypergraph Laplacian regularization (HLapR) and $p$-Laplacian regularization (pLapR). The contributions of this paper can be summarized as below.

1) We present an efficient approximation algorithm of Hypergraph $p$-Laplacian, significantly improving computation efficiency.
2) We propose HpLapR to preserve the local similarity of data.
3) We integrate HpLapR into logistic regression and conduct comprehensive experiments to empirically analyze our method on UC-Merced data set. The experimental results validate the effectiveness of our method.

The rest of the paper is organized as follows. Section 2 briefly reviews related work on manifold regularization and hypergraph learning. Section 3 introduces the proposed HpLapR including an approximate computation of the Hypergraph $p$-Laplacian. Section 4 presents the HpLapR logistic regression. Section 5 presents the experimental results and analysis on UC-Merced data set. Finally, Section 6 gives the conclusions.

## II. RELATED WORK

In this section, we briefly review related works of manifold regularization and hypergraph.

### A. Manifold Regularization

Assume the estimated function, which is generated from the probability distribution on examples (labeled examples and unlabeled examples). The labeled examples are $(x, y)$ pairs generated according to probability distribution. And supposing the labeled examples lied on the estimation curve in the ideal case. The unlabeled examples are drawn according to the marginal distribution. Based on the manifold assumption that if two examples are close in the intrinsic geometry, then the two examples have the similar labels, it is important to exploit the knowledge of the marginal distribution for better function learning.

By introducing an additional regularizer for local structure preserving, the manifold regularization framework can be interpreted as regularization algorithms with different empirical cost functions, complexity measures in an appropriately chosen Reproducing Kernel Hilbert Space (RKHS) and additional

information about the geometric structure of the marginal. Hence, the objective function can be written as

$$f^* = arg\,min_{f \in H_K} \frac{1}{l} \sum_{i=1}^{l} V(x_i, y_i, f) + Y_A \|f\|_K^2 + Y_I \|f\|_I^2. \quad (1)$$

Where $V$ is some loss function, such as the hinge loss function $max\,[0, 1 - y_i f(x_i)]$ for Support Vector Machines (SVM). The corresponding norm ($\|f\|_K^2$) is used to control the complexity of the classification model, while $\|f\|_I^2$ is an appropriate penalty term corresponding to the probability distribution. The parameters $Y_A$ and $Y_I$ control the complexity of the function in the ambient space and the intrinsic geometry, respectively.

Graph Laplacian has been widely used to explore and exploit the local geometry of data distribution. As a nonlinear generalization of the standard graph Laplacian, graph $p$-Laplacian has attracted attentions from machine learning community. Zhou and Schölkopf [34] proposed a general discrete regularization framework of $p$-Laplacian for the classification problem, and its objective function can be computed as follows:

$$f^* = argmin_{f \in \mathcal{H}(V)} \{\mathcal{S}_p(f) + \mu \|f - y\|^2\} \quad (2)$$

where $\mathcal{S}_p(f) := \frac{1}{2} \sum_{v \in V} \|\nabla_v f\|^p$ is the $p$-Dirichlet form of the function $f$, $\mu$ is a parameter balancing the two competing terms, $y \in \{-1, 0, 1\}$ depends on the label of example.

Bühler and Hein [32] used the graph $p$-Laplacian for spectral clustering and demonstrated the relationship between the second eigenvalue of the graph $p$-Laplacian and the optimal Cheeger cut as follows:

$$RCC \le RCC^* \le p(\frac{max}{i \in V}d_i)^{\frac{p-1}{p}} RCC^{\frac{1}{p}} \quad (3)$$

or

$$NCC \le NCC^* \le pNCC^{\frac{1}{p}} \quad (4)$$

where $RCC^*$ and $NCC^*$ as the ratio/normalized Cheeger cut values obtained by tresholding the second eigenvector of the unnormalized/normalized $p$-Laplacian, $d_i$ is the degree of vertex $i$, $RCC$ and $NCC$ as the optimal ratio /normalized Cheeger cut.

Luo *et al.* [35] used the $p$-Laplacian for multi-class clustering and provided an approximation of the whole eigenvectors by solving the tractable optimization problem:

$$\min_{\mathcal{F}} J_E(\mathcal{F}) = \sum_k \frac{\sum_{ij} w_{ij} |f_i^k - f_j^k|^p}{\|f^k\|_p^p}$$
$$s.t. \quad \mathcal{F}^T \mathcal{F} = I. \quad (5)$$

Where $w_{ij}$ is the edge weight, $f^k$ is an eigenvector of $p$-Laplacian, $\mathcal{F} = (f^1, f^2, \cdots, f^n)$ are whole eigenvectors.

Liu *et al.* [36] proposed $p$-Laplacian regularized sparse coding for preserving the manifold structure.

### B. Hypergraph

In machine learning issues, we generally assume pairwise relationships among the objects set. An object set endowed with pairwise relationships can be considered as a graph. The graph can be undirected or directed. However, in a number of questions, it is not complete to represent the relations among samples only by simple graphs. Hypergraph learning [22] addresses the problem. Comparing with traditional graph, a hypergraph illustrates the complex relationships by hyperedges which connect three or more vertices (see in Fig. 1).

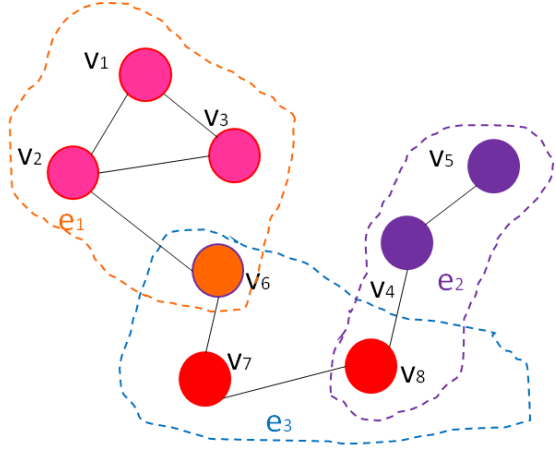

| | e₁ | e₂ | e₃ |
|---|---|---|---|
| V₁ | 1 | 0 | 0 |
| V₂ | 1 | 0 | 0 |
| V₃ | 1 | 0 | 0 |
| V₄ | 0 | 1 | 0 |
| V₅ | 0 | 1 | 0 |
| V₆ | 1 | 0 | 1 |
| V₇ | 0 | 0 | 1 |
| V₈ | 0 | 1 | 1 |

Fig. 1. The block scheme of hypergraph. Left: A simple graph in which two points are joined together by an edge if they are highly similarity. A hypergraph completely illustrates the complex relationships among points by hyperedges. Right: The $H$ matrix of the hypergraph. The entry $(v_i, e_j)$ is set to 1 if a hyperedge $e_j$ contains $v_i$, or 0 otherwise.

Let $V$ denote a finite set of vertices and $E$ a family of subsets of $V$ such that $\bigcup_{e \in E} = V$. A hypergraph $G = (V, E)$ corresponding to the vertex set $V$ and the hyperedge set $E$. Denote the weight associated with each hyperedge $e$ as $w(e)$. The degree of a vertex $v \in V$ is defined by $d(v) = \sum_{\{e \in E \mid v \in e\}} w(e)$. The degree of a hyperedge $e \in E$ is denoted as $\delta(e) = |e|$. Denote the incident matrix $H$ by a $|V| \times |E|$ matrix, whose entry $h(v, e) = 1$ if $v \in e$, and $h(v, e) = 0$ otherwise. Then $d(v) = \sum_{e \in E} w(e) h(v, e)$, $\delta(e) = \sum_{v \in V} h(v, e)$. Let $Dv$ denote the diagonal matrices containing the degree of vertex, $D_e$ denote the diagonal degree matrices of each hyperedge, and $W$ is the diagonal matrix of edge weights. Then, the hypergraph Laplacian can be defined.

There have been many methods for building the graph Laplacian of hypergraphs across literature. The first category includes star expansion [28], clique expansion [28], Rodriguez's Laplacian [29], etc. These methods aim to construct a simple graph from the original hypergraph, and then partitioning the vertices by spectral clustering techniques. The second category of approaches defines a Hypergraph Laplacian using analogies from the simple graph Laplacian. Representative methods in this category include Bolla's Laplacian [30], Zhou's normalized Laplacian [31], etc. In [31], the normalized Hypergraph Laplacian is defined as

$$L^{hp} = I - Dv^{-1/2} H W D_e^{-1} H^T Dv^{-1/2}. \quad (6)$$

Note that $L^{hp}$ is positive semi-definite. The adjacency matrix of hypergraph can be formulated as follows:

$$W^{hp} = HWH^T - Dv. \quad (7)$$

For a simple graph, the edge degree matrix $D_e$ is replaced to $2 I$. Thus, the standard graph Laplacian is

$$L = I - \frac{1}{2} Dv^{-\frac{1}{2}} H W H^T Dv^{-\frac{1}{2}}$$
$$= \frac{1}{2} (I - Dv^{-1/2} W^{hp} Dv^{-1/2}). \quad (8)$$

## III. HPLAPR

In SSL, we are given $N$ training samples including $l$ labeled samples $\{(x_i, y_i)\}_{i=1}^l$ and $u$ unlabeled samples $\{(x_j)\}_{j=l+1}^{l+u}$.

Class labels are given in $Y = \{y_i\}_{i=1}^l$, where $y_i \in \{\pm 1\}$. Typically, $l \ll u$ and the goal is to predict the labels of unseen examples.

According to the manifold regularization framework, the proposed HpLapR can be written as the following optimization problem:

$$f^* = arg\ min_{f \in H_K} \frac{1}{l} \sum_{i=1}^l V(x_i, y_i, f) + Y_A \|f\|_K^2$$
$$+ \frac{Y_I}{(l+u)^2} \mathbf{f}^T L_p^{hp} \mathbf{f}. \quad (9)$$

Here, $\mathbf{f}$ is given as $\mathbf{f} = [f(x_1), f(x_2), \cdots, f(x_{l+u})]^T$, $L_p^{hp}$ is the Hypergraph $p$-Laplacian.

### A. Approximation of Hypergraph $p$-Laplacian

In this sub-section, we briefly describe the approximating of Hypergraph $p$-Laplacian $L_p^{hp}$.

Assume that the Hypergraph $p$-Laplacian has $n$ eigenvectors $\mathcal{F}^{*hp} = (f^{*hp1}, f^{*hp2}, \cdots, f^{*hpn})$ associated with unique eigenvalues $\lambda^{*hp} = (\lambda_1^{*hp}, \lambda_2^{*hp}, \cdots, \lambda_n^{*hp})$, we compute the approximation of $L_p^{hp}$ by $L_p^{hp} = \mathcal{F}^{*hp} \lambda^{*hp} \mathcal{F}^{*hp^T}$. Thus, it is important to get all eigenvectors and eigenvalues of Hypergraph $p$-Laplacian.

Although a complete analysis of Hypergraph $p$-Laplacian is challenging, we can easily generate a hypergraph with a group of hyperedges [31]. In details, we construct the hypergraph Laplacian $L^{hp}$ and compute adjacency matrix $W^{hp}$ by (6) and (7), respectively.

Then, we introduce the basic definition of $p$-Laplacian $\Delta_p^w$ including eigenvalue and eigenvector.

The real number $\lambda_p$ is called as an eigenvalue for $\Delta_p^w$, if there exists a function $f : V \to \mathcal{R}$ satisfying the relationship as following:

$$(\Delta_p^w f)_i = \lambda_p \phi_p(f_i), \ i \in V. \quad (10)$$

The function $f$ is called a $p$-eigenfunction (also called eigenvector) associated with $\lambda_p$. Where $\phi_p$ is defined by $\phi_p(x) = |x|^{p-1} sign(x)$. Note that the operator $\Delta_2^w = L$ becomes the regular graph Laplacian.

Following previous studies on $p$-Laplacian [32], the computation of eigenvalue and the corresponding eigenvector on nonlinear operator $\Delta_p^w$ can be solved by the theorem:

The functional $F_p$ has a critical point at $f$ if and only if $f$ is an eigenvector of $\Delta_p^w$. $F_p$ is defined as:

$$F_p(f) = \frac{\Sigma_{ij} w_{ij} |f_i - f_j|^p}{2\|f\|_p^p} \quad (11)$$

where

$$\| f \|_p^p = \Sigma_i |f_i|^p.$$

Here $w_{ij}$ is the edge weight, the corresponding eigenvalue $\lambda_p$ is given by $\lambda_p = F_p(f)$. The above theorem serves as the foundational analysis of eigenvectors and eigenvalues. Moreover $F_p(\alpha f) = F_p(f)$ apply for all real value $\alpha$.

Naturally, we can extend the above theorem to the Hypergraph $p$-Laplacian as follows:

$f^{hp}$ is an eigenvector of hypergraph $p$-Laplacian, if and only if the following function $F_p^{hp}$ has a critical point at $f^{hp}$:

$$F_p^{hp}(f^{hp}) = \frac{\Sigma_{ij} w_{ij}^{hp} |f_i^{hp} - f_j^{hp}|^p}{2\|f_i^{hp}\|_p^p} \quad (12)$$

where

$$\| f^{hp} \|_p^p = \Sigma_i |f_i^{hp}|^p.$$

The eigenvalue $\lambda^{hp}$ associated with $f^{hp}$ is given by $\lambda^{hp} = F_p^{hp}(f^{hp})$.

If we want to get all eigenvectors and eigenvalues of hypergraph $p$-Laplacian, we have to find all critical points of the function $F_p^{hp}$. Following this idea, we can get the full eigenvectors space by solving local solution of the following optimization problem:

$$\min_{\mathcal{F}^{hp}} J(\mathcal{F}^{hp}) = \Sigma_k F_p^{hp}(f^{hpk})$$
$$s.t. \ \Sigma_i \phi_p(f_i^{hpk}) \phi_p(f_i^{hpl}) = 0, k \neq l \quad (13)$$

where $\mathcal{F}^{hp} = (f^{hp1}, f^{hp2}, \cdots, f^{hpn})$.

We analyze the full eigenvectors by solving the following Hypergraph $p$-Laplacian embedding problem instead of (13):

$$\min_{\mathcal{F}^{hp}} J_E(\mathcal{F}^{hp}) = \sum_k \frac{\Sigma_{ij} w_{ij}^{hp} |f_i^{hp} - f_j^{hp}|^p}{\| f^{hp} \|_p^p}$$
$$s.t. \ \mathcal{F}^{hp^T} \mathcal{F}^{hp} = I. \quad (14)$$

Differentiating with respect to $f_i^{hpk}$ yields the following equation:

$$\frac{\partial J_E}{\partial f_i^{hpk}} = \frac{1}{\|f^{hpk}\|_p^p} \left[ \Sigma_j w_{ij}^{hp} \phi_p(f_i^{hpk} - f_j^{hpk}) - \frac{\phi_p(f_i^{hpk})}{\|f^{hpk}\|_p^p} \right]. \quad (15)$$

Solving the problem (14) with the gradient descend optimization, the gradient is defined in the following way:

$$G^{hp} = \frac{\partial J_E}{\partial \mathcal{F}^{hp}} - \mathcal{F}^{hp} \left( \frac{\partial J_E}{\partial \mathcal{F}^{hp}} \right)^T \mathcal{F}^{hp}. \quad (16)$$

Meanwhile, the full eigenvalue $\lambda^{hp} = (\lambda_1^{hp}, \lambda_2^{hp}, \cdots, \lambda_n^{hp})$ can be computed by $\lambda_k^{hp} = \frac{\Sigma_{ij} w_{ij}^{hp} |f_i^{hp} - f_j^{hpk}|^p}{\|f^{hpk}\|_p^p}$.

Finally, the approximation of $L_p^{hp}$ can be solved by the full eigenvectors and eigenvalues of hypergraph $p$-Laplacian in this paper. We summarize the approximation of Hypergraph $p$-Laplacian in Algorithm 1. In the algorithm, the step length $\alpha$ is set to be $\alpha = 0.01 \frac{\Sigma_{ik} |\mathcal{F}_{ik}^{hp}|}{\Sigma_{ik} |G_{ik}^{hp}|}$.

---

**Algorithm 1** The Approximate of Hypergraph $p$-Laplacian

**Input:** Training examples $X$; Embedding dimension $K$; $p$

**output:** hypergraph $p$-Laplacian: $L_p^{hp}$

Step1: Construct hypergraph Laplacian matrix $L^{hp}$ and compute data adjacency matrix $W^{hp}$.

Step 2: Decomposition of graph Lapalcian: $L^{hp} = USU^T$.

**Initialize:** $\mathcal{F}^{hp} = U(:, 1:K)$

Step 3: **While not converged do:**

$G^{hp} = \frac{\partial J_E}{\partial \mathcal{F}^{hp}} - \mathcal{F}^{hp} \left( \frac{\partial J_E}{\partial \mathcal{F}^{hp}} \right)^T \mathcal{F}^{hp}$, where $\frac{\partial J_E}{\partial \mathcal{F}^{hp}}$ is given by Equation (14)

$\mathcal{F}^{hp} = \mathcal{F}^{hp} - \alpha G^{hp}$

**End**

Step 4: $\lambda_k^{hp} = \frac{\Sigma_{ij} w_{ij}^{hp} |f_i^{hpk} - f_j^{hpk}|^p}{\|f^{hpk}\|_p^p}$

**return:** $L_p^{hp} = \mathcal{F}^{hp} \lambda^{hp} \mathcal{F}^{hp^T}$

---

## IV. HpLapR Logistic Regression

Actually, the proposed HpLapR can be applied to variant applications by integrating different choices of loss function $V(x_i, y_i, f)$ into manifold regularization framework. In this section, we apply the HpLapR to logistic regression and give the complexity analysis.

Substitute logistic loss function into (9), the HpLapR can be rewritten as

$$f^* = arg \min_{f \in H_K} \frac{1}{l} \sum_{i=1}^{l} \left( log(1 + e^{-y_i f(x_i)}) \right) + \Upsilon_A \|f\|_K^2$$
$$+ \frac{\Upsilon_I}{(l+u)^2} \mathbf{f}^T L_p^{hp} \mathbf{f}. \quad (17)$$

The classical Representer Theorem indicates the solution of (17) w.r.t. $f$ exists and can be expressed as

$$f^*(x) = \sum_{i=1}^{l+u} \alpha_i^* \boldsymbol{K}(x_i, x). \quad (18)$$

The ambient kernel $\boldsymbol{K}$ is symmetric positive definite. Thus, we finally construct the HpLapR as the following optimization problem:

$$f^* = arg \min_{f \in H_K} \frac{1}{l} \sum_{i=1}^{l} \left( log(1 + e^{-y_i K(x_i, x)\alpha}) \right)$$
$$+ \Upsilon_A \alpha^T \boldsymbol{K}\alpha + \frac{\Upsilon_I}{(l+u)^2} \alpha^T \boldsymbol{K} L_p^{hp} \boldsymbol{K}\alpha. \quad (19)$$

To solve the optimization problem in (19), we can employ the conjugate gradient algorithm. We take derivative of the objective function as

$$\nabla f(\alpha) = -\frac{log(e)}{l} \sum_{i=1}^{l} \left( \frac{y_i}{1 + e^{y_i K(x_i, x)\alpha}} \boldsymbol{K}^T(x_i, x) \right)$$
$$+ \Upsilon_A (\boldsymbol{K} + \boldsymbol{K}^T)\alpha + \frac{\Upsilon_I}{(l+u)^2} \left( \boldsymbol{K} L_p^{hp} \boldsymbol{K} + \left( \boldsymbol{K} L_p^{hp} \boldsymbol{K} \right)^T \right) \alpha. \quad (20)$$

The optimization procedure of conjugate gradient algorithm for HpLapR logistic regression is described in Algorithm 2.

Suppose we are given $N$ samples. Denote the embedding dimension as $K$, and the number of iteration as $\eta_1$ for approximating of Hypergraph $p$-Laplacian. The time cost for constructing Hypergraph $p$-Laplacian is $O(\eta_1(N^2 K + nK^2))$. When $K$ is much smaller than $N$, the time cost is around $O(\eta_1 N^2)$. Denote the number of iterations as $\eta_2$ for HpLapR logistic regression and the number of candidate parameters that need the m-fold cross-validation as $r$. The time cost for HpLapR logistic regression is $O(\eta_2 r N^3)$.

---

**Algorithm 2** HpLapR Logistic Regression

**Input:** $l$ labeled samples $\{(x_i, y_i)\}_{i=1}^l$,
$u$ unlabeled samples $\{(x_j)\}_{j=l+1}^{l+u}$.
**output:** Estimated function: $f^*(x) = \sum_{i=1}^n \alpha_i^* K(x_i, x)$.
Step1: Construct approximate Hypergraph $p$-Laplacian $L_p^{hp}$.
Step2: Choose a kernel function and compute the Gram matrix $K_{ij} = K(x_i, x_j)$.
Step3: Compute $\alpha^*$:
**Initialize:** $\alpha^0 \in R^N$, $d^0 = -\nabla f(\alpha)$, $\delta, 0 < \varepsilon \ll 1$, $m = 0$
while $|f(\alpha^{m+1}) - f(\alpha^m)| > \varepsilon$
**do:**
$\alpha^{m+1} = \alpha^m + \delta d^m$
$d^{m+1} = -\nabla f(\alpha^{m+1}) + \frac{\|\nabla f(\alpha^{m+1})\|^2}{\|\nabla f(\alpha^m)\|^2} d^m$
$m = m + 1$
**return:** $\alpha^* = \alpha^{m+1}$

---

## V. EXPERIMENTS

In this section, we will evaluate the effectiveness of the proposed HpLapR by compared with other local structure preserving algorithms including LapR, HLapR and pLapR. We apply the logistic regression for remote sensing image classification. Fig. 2 illustrates the framework of HpLapR for UC-Merced data set.

UC-Merced data set [25] consists of totally 2100 land-use images collected from aerial orthoimage with the pixel resolution of one foot. The original images were downloaded from the United States Geological Survey National Map of 20 U.S. regions. These images were manually selected into 21 classes: agricultural, airplane, baseball diamond, beach, buildings, chaparral, dense residential, forest, freeway, golf course, harbor, intersection, medium density residential, mobile home park, overpass, parking lot, river, runway, sparse residential, storage tanks, and tennis courts. In this paper, we organized these 21 classes into six groups (see in Fig. 3). Note that UC-Merced data set contains a variety of land-use classes, which make the data set more challenging. Specially, some highly overlapped classes, *e.g.*, sparse residential, medium density residential, and dense residential that mainly differ in the density, make it a difficult classification task.

In our experiments, we extract high-level visual features using the deep convolution neural network (CNN) [26]. We randomly choose the 50 images per class as training samples and the rest as testing samples. For hypergraph construction, we regard each sample in the training set as a vertex, and generate a hyperedge for each vertex with its $k$ nearest neighbors (so the hyperedge connects $k + 1$ samples) [33]. It is worth noticing that, for our experiments, the $k$ NN-based hyperedges generating method is not implemented in the overall training samples, but in six groups. For example, for a sample of baseball diamond, the vertices of the corresponding hyperedge are just chosen from the first group (baseball diamond, golf course and tennis courts) of Fig. 3.

In semi-supervised classification experiments, we assign 10%, 20%, 30%, 50% samples of training data as labeled data, while the rest are used as unlabeled data. The process is repeated for five times independently to avoid any bias introduced by the random partitioning of data.

We conduct the experiments on our data set to get the proper modal parameters. The neighborhood size $k$ of a hypergraph varies in a range $\{5, 6, 7, \cdots, 15\}$ through cross-validation. The regularization parameters $\gamma_A$ and $\gamma_I$ are selected from the candidate set $\{10^i | i = -10, -9, -8, \cdots, 10\}$ through cross-validation, and the parameter $p$ for pLapR and HpLapR are chosen from $\{1, 1.1, 1.2, \cdots, 3\}$ through cross-validation with 10% labeled samples on the training data, respectively. We verify the classified performance by average precision (AP) performance for single class and mean average precision (mAP) [27] for overall classes.

Fig. 4 illustrates the mAP performance of pLapR and HpLapR on the validation set when $p$ varies. The $x$-axis is the parameter $p$ and the $y$-axis is mAP for performance measure. We can see that the best mAP performance for pLapR can be obtained with $p = 2.3$ while best performance for the HpLapR is achieved when $p$ is equal to 2.6.

We compare our proposed HpLapR with the representative LapR, HLapR and pLapR. From Fig. 5, we can observe that, HpLapR usually outperforms other methods especially when only a small number of samples labeled. This suggests that our proposed method which considering the hypergraph learning and $p$-Laplacian has the superiority to preserve the local structure of the data.

To evaluate the effectiveness of HpLapR for single class, Fig. 6 shows the AP results of different methods on several selected land-use classes including beach, dense residential, freeway and tennis court. From Fig. 6, we can find that, in most cases, the HpLapR performs better than both pLapR and HLapR, while pLapR and HLapR consistently outperforms than LapR.

## VI. CONCLUSION

The existing successful SSL algorithms have achieved great performance in computer vision applications including classification, clustering, ranking, etc. However, how to obtain the high-order relationship and exploit the local geometry of the data distribution is still challenging. Therefore, we present a Hypergraph $p$-Laplacian regularized method to preserve the geometry of the probability distribution. Both hypergraph and $p$-Laplacian have the advantage in local structure preserving. Furthermore, we introduced a fully approximation algorithm of Hypergraph $p$-Laplacian lowing down its computation difficulties. Finally, we propose Hypergraph $p$-Laplacian regularized logistic regression for remote sensing recognition. We present the experimental results on UC-Merced dataset to demonstrate the efficacy of our proposed method in comparison to other regularized methods including LapR, HLapR and pLapR.

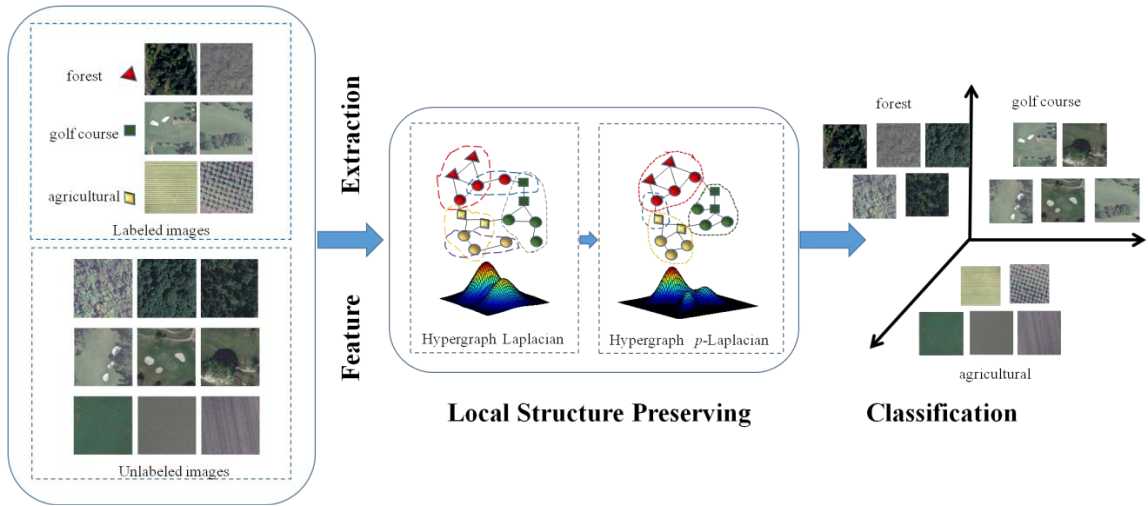

Fig. 2. The framework of HpLapR for remote sensing image classification.

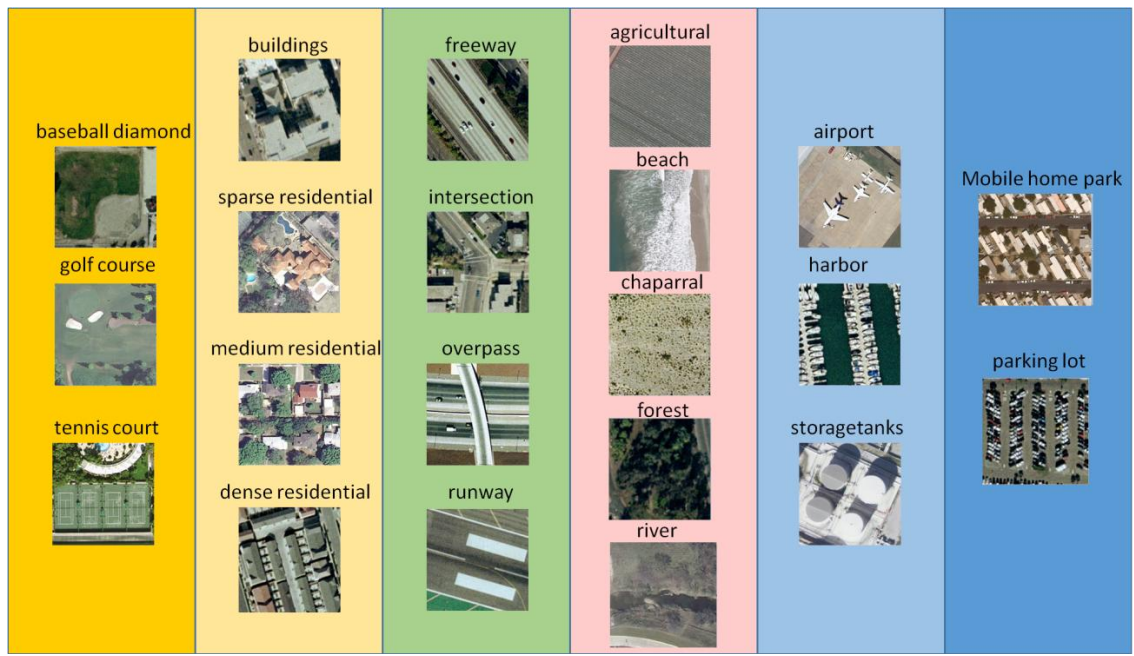

Fig. 3. Some examples of UC-Merced data set. The dataset totally has 21 remote sensing categories that can be simply grouped into six groups according to the distinction of land use. Each column represents one group.

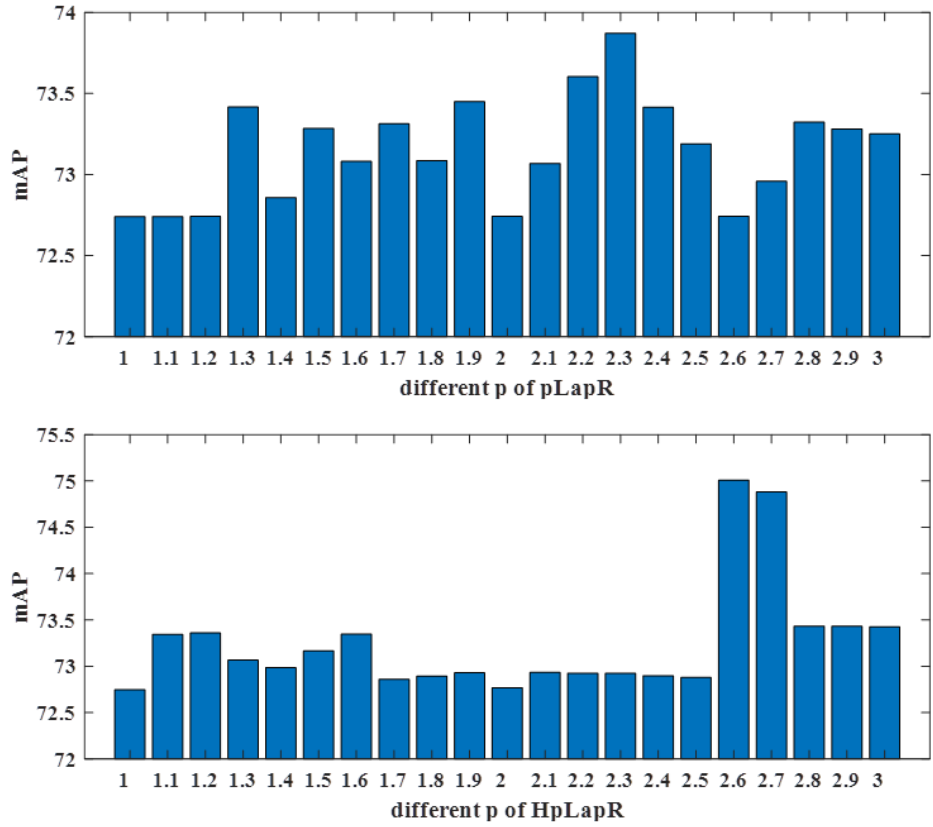

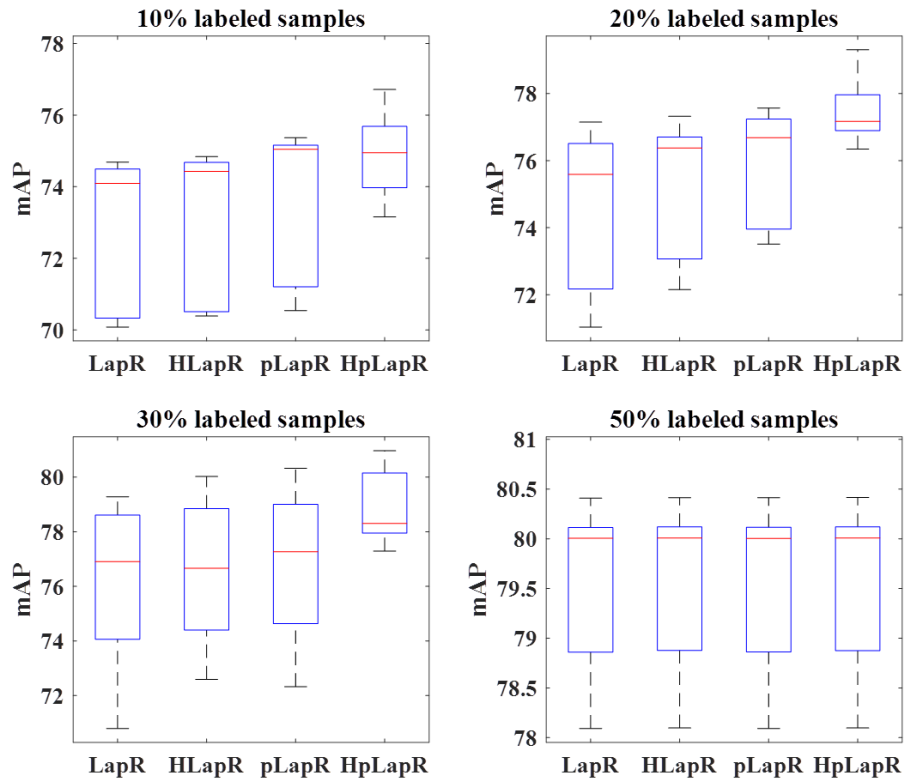

Fig. 4. Performance of mAP with different $p$ on validation set.

Fig. 5. mAP performance of different algorithms.

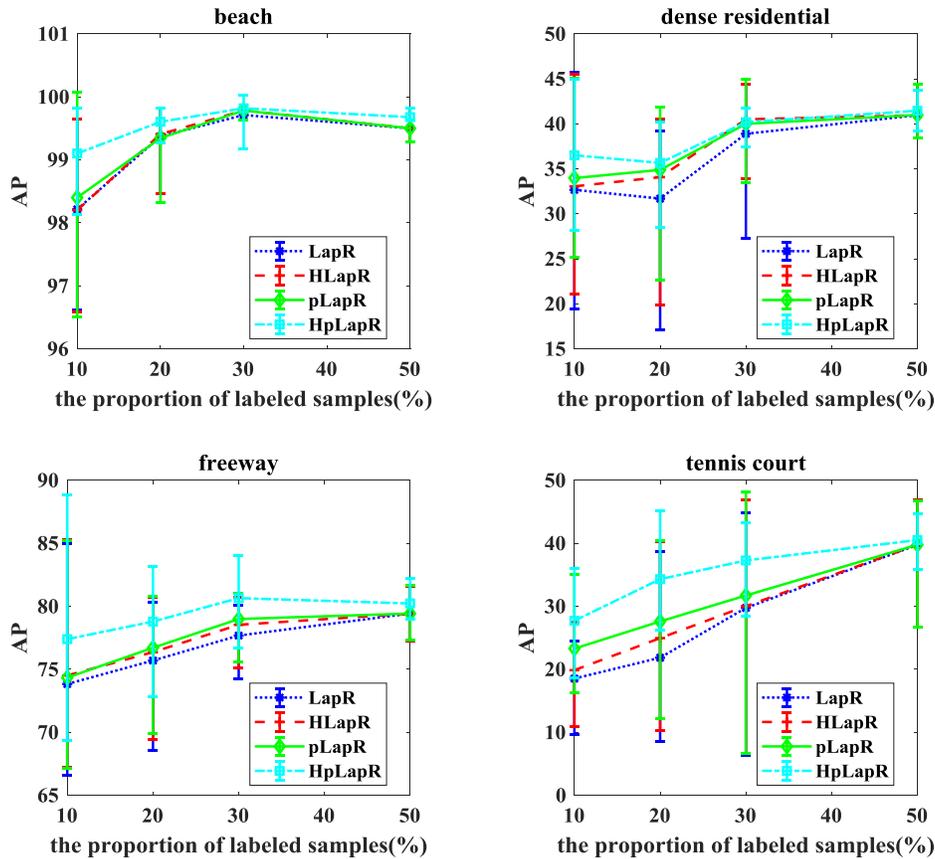

Fig. 6. AP performance of different methods on some classes including beach, dense residential, freeway and tennis court.

**Xueqi Ma** is perusing her master degree with the College of Information and Control Engineering, China University of Petroleum (East China), China. Her research interests include pattern recognition, computer vision.

**Weifeng Liu** is currently a full Professor with the College of Information and Control Engineering, China University of Petroleum (East China), China. He received the double B.S. degree in automation and business administration and the Ph.D. degree in pattern recognition and intelligent systems from the University of Science and Technology of China, Hefei, China, in 2002 and 2007, respectively. His current research interests include computer vision, pattern recognition, and machine learning. inventions.

**Yichong Zhou** received the B.S. degree from Hunan University, Changsha, China, and the M.S. and Ph.D. degrees from Tufts University, MA, USA, all in electrical engineering. He is currently an Associate Professor and the Director of the Vision and Image Processing Laboratory, Department of Computer and Information Science, University of Macau, Macau, China. His research interests include chaotic systems, multimedia security, image processing and understanding, and machine learning.